\documentclass[11pt]{article}

\usepackage[]{acl}

\usepackage{times}
\usepackage{latexsym}

\usepackage[T1]{fontenc}

\usepackage[utf8]{inputenc}

\usepackage{microtype}

\usepackage{inconsolata}

\usepackage{graphicx}

\usepackage{bbding} 
\usepackage{pifont}
\usepackage{xcolor}
\usepackage{booktabs}
\usepackage{url}
\usepackage{multirow}
\usepackage{makecell}
\usepackage{arydshln}
\usepackage{amsmath}
\usepackage{fontawesome5}

\title{KDFlow: A User-Friendly and Efficient Knowledge Distillation \\ Framework for Large Language Models}

\author{Songming Zhang\textsuperscript{1,2,3}\thanks{Corresponding author.}, Xue Zhang\textsuperscript{1,2}, Tong Zhang\textsuperscript{3}, Bojie Hu\textsuperscript{3}, \\
\textbf{Yufeng Chen}\textsuperscript{1,2}, and 
\textbf{Jinan Xu}\textsuperscript{1,2} \\
\textsuperscript{1} School of Computer Science and Technology, Beijing Jiaotong University, Beijing, China \\
\textsuperscript{2} Key Laboratory of Big Data \& Artificial Intelligence in Transportation,\\(Beijing Jiaotong University), Ministry of Education \\
\textsuperscript{3} Weixin AI \\
\texttt{\{smzhang22,zhang\_xue,chenyf,jaxu\}@bjtu.edu.cn}}

\begin{document}
\maketitle
\begin{abstract}
Knowledge distillation (KD) is widely used to compress and post-train large language models (LLMs), yet many existing frameworks execute teacher inference with the same training-oriented backend as student optimization, leading to suboptimal efficiency.
In this paper, we propose \textbf{KDFlow}, a novel framework for LLM distillation that features a \textbf{decoupled architecture} and employs SGLang for teacher inference.
KDFlow combines SGLang for teacher inference with PyTorch FSDP2 for student optimization, allowing each model to run on a backend tailored to its workload.
To enable \textbf{efficient full-vocabulary distillation} in this decoupled architecture, KDFlow transfers the teacher’s final hidden states via Ray's object store and recomputes teacher logits on each student worker using a frozen copy of the teacher's output head.
Furthermore, our framework supports both off-policy and on-policy distillation and incorporates cross-tokenizer algorithms through highly extensible and user-friendly APIs. 
Experiments show that KDFlow achieves a \textbf{1.44$\times$ to 6.36$\times$} speedup over MS-SWIFT in off-policy distillation and a \textbf{1.43$\times$ to 1.75$\times$} speedup over verl in on-policy distillation.
KDFlow further scales to 64 GPUs, achieving 3.68$\times$ and 2.52$\times$ strong-scaling speedups in two representative model configurations.
The code and documentation are publicly available.
\end{abstract}

\begin{center}
    \small
    \href{https://github.com/songmzhang/KDFlow}{\faGithub\ \textbf{Code}} \quad
    \href{https://kdflow.readthedocs.io/}{\faBook\ \textbf{Documentation}}
\end{center}

\section{Introduction}

Large Language Models (LLMs) have demonstrated remarkable capabilities across diverse tasks, yet their substantial parameter scales pose significant challenges for deployment in resource-constrained environments.
Knowledge Distillation (KD) offers an effective approach to addressing this challenge by transferring knowledge from a large teacher model to a compact student model \cite{hinton2015distilling}.

\begin{figure}
    \centering
    \includegraphics[width=\linewidth]{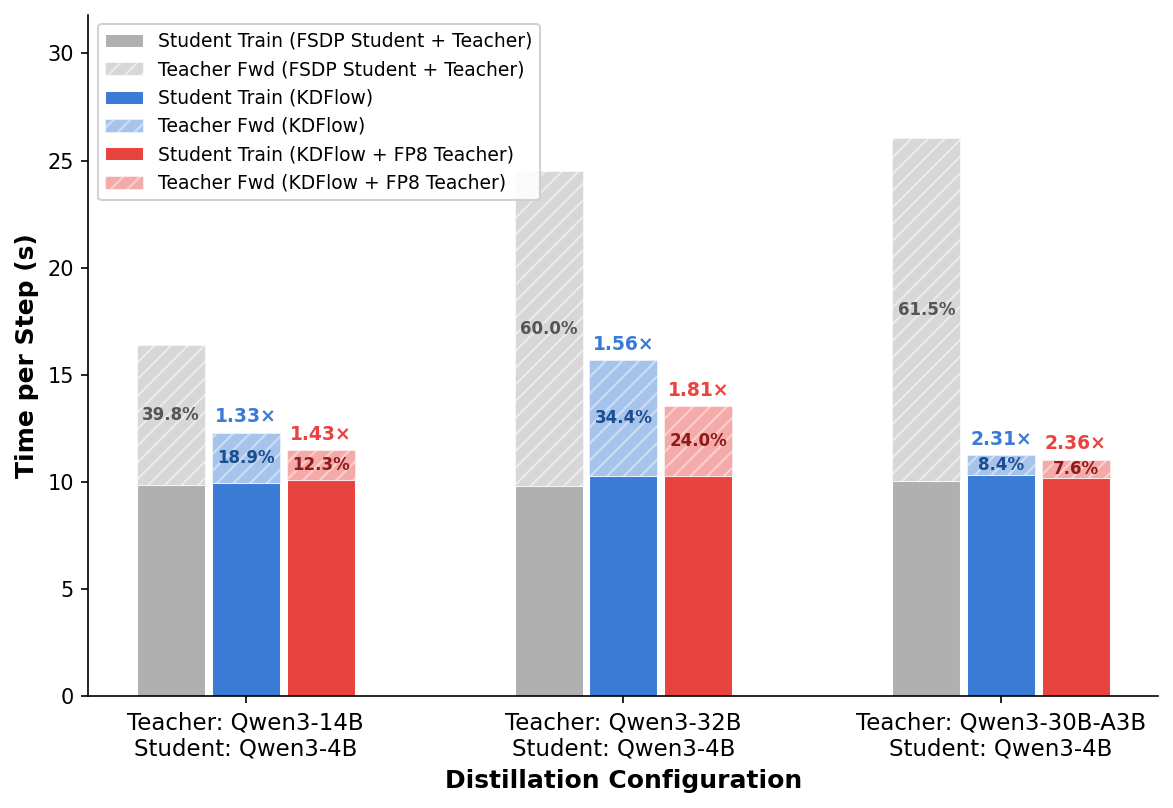}
    \caption{Training time per step and the proportion of teacher forward time under different distillation configurations. The teacher's MoE architecture poses challenges for FSDP, while being well-supported by SGLang.}
    \label{fig:tea_fwd_occ}
\end{figure}

Despite extensive research and application of KD, the infrastructure for LLM distillation remains suboptimal. 
In a typical KD process, the teacher and student models play distinct roles: the teacher performs only forward passes (inference), while the student requires both forward and backward passes (training).
However, most existing frameworks, such as TRL \cite{vonwerra2020trl} or MS-SWIFT \cite{zhao2025swift}, run both models with a unified training backend.
This creates a structural mismatch, as a training engine designed for gradient computation and optimizer state management is ill-suited to the teacher's inference-heavy workload. 
As a result, the throughput of LLM distillation is often bounded by inefficient teacher execution, especially for Mixture-of-Experts (MoE) teachers (see Figure \ref{fig:tea_fwd_occ}).

\begin{table*}[t]
    \centering
    \setlength{\tabcolsep}{4pt}
    \renewcommand{\arraystretch}{1.15}
    \resizebox{\linewidth}{!}{
    \begin{tabular}{lccccccc}
        \toprule
        \textbf{Features} & \textbf{TRL} & \textbf{MS-SWIFT} & \textbf{EasyDistill} & \textbf{ROLL} & \textbf{Slime} & \textbf{verl} & \textbf{KDFlow (Ours)} \\
        \midrule
        \textbf{KD-Native Design} & & & \Checkmark & & & & \Checkmark \\
        \textbf{Decoupled Backends} & & \Checkmark & & & \Checkmark & \Checkmark & \Checkmark \\
        \textbf{Off-Policy Distillation} & \Checkmark & \Checkmark & \Checkmark & \Checkmark & & & \Checkmark \\
        \textbf{On-Policy Distillation} & \Checkmark & \Checkmark & & \Checkmark & \Checkmark & \Checkmark & \Checkmark \\
        \textbf{Self Distillation} & & \Checkmark & & & & \Checkmark & \Checkmark \\
        \textbf{Cross-Tokenizer Distillation} & \Checkmark & & & & & & \Checkmark \\
        \textbf{Multi-Teacher Distillation} & & \Checkmark & & \Checkmark & \Checkmark & \Checkmark & \Checkmark \\
        \midrule
        \textbf{Logits/Logprobs} & \textbf{full vocab} & \textbf{full vocab} & \textbf{full vocab} & \makecell{top-$k$ vocab\\sampled tokens} & sampled tokens & \makecell{top-$k$ vocab\\sampled tokens} & \textbf{full vocab} \\
        \textbf{Divergence Metrics} & \makecell{FKL,\\RKL,\\JSD} & \makecell{FKL,\\RKL,\\JSD} & \makecell{FKL,\\RKL} & \makecell{\textit{off-policy}: FKL, RKL, JSD, \\Skewed F(R)KL, AKL, \\\textit{on-policy}: RKL} & RKL & RKL & \makecell{\textbf{FKL, RKL, JSD, }\\\textbf{Skewed F(R)KL, }\\\textbf{AKL, TVD}} \\
        \bottomrule
        
    \end{tabular}
    }
    \caption{Comparisons between KDFlow and existing frameworks.}
    \label{tab:framework_compare}
\end{table*}

To address this issue, we present \textbf{KDFlow}, a high-performance framework designed specifically for LLM distillation. 
KDFlow decouples the teacher and student backends by assigning the student to PyTorch FSDP2 and the teacher to the high-throughput inference engine SGLang \cite{zheng2024sglang}.
A key challenge in this architecture is transferring full teacher logits from SGLang processes to FSDP processes: directly transferring full-vocabulary logits is prohibitively expensive for current LLM training\footnote{For 128 sequences with length 4096, full BF16 logits from Qwen3 models occupy $128 \times 4096 \times 151936 \times 2\text{bytes}\approx160\text{GB}$ of memory.}, while transferring only top-k logits breaks the mathematical equivalence of the distillation objective.
KDFlow solves this by collecting compact teacher hidden states from SGLang and recomputing the full logit distribution on the student side, reducing communication overhead while preserving standard KD.
Compared with MS-SWIFT, KDFlow delivers a \textbf{1.44$\times$ to 6.36$\times$} training speedup in off-policy distillation, with larger gains for Mixture-of-Experts (MoE) teachers.
In end-to-end on-policy distillation, KDFlow achieves a 1.43$\times$ to 1.75$\times$ speedup over verl and demonstrates effective scaling from 8 to 64 GPUs.

Despite the decoupled design, KDFlow abstracts away distributed communication and integrates with standard Hugging Face model formats, allowing users to initiate KD with only a few lines of configuration.

Overall, the contributions of KDFlow include:
\begin{itemize}
    \item \textbf{Efficient Architecture}: KDFlow decouples teacher inference from student training by serving the teacher with SGLang and transferring compact hidden states across processes, achieving a 1.44$\times$ to 6.36$\times$ speedup over MS-SWIFT in off-policy distillation, a 1.43$\times$ to 1.75$\times$ speedup over verl in on-policy distillation, and effective scaling to 64 GPUs.
    \item \textbf{Comprehensiveness}: KDFlow supports comprehensive KD features (e.g., off/on-policy, cross-tokenizer, and multi-teacher distillation) and provides multiple built-in divergence metrics and algorithms, making it an out-of-the-box toolkit for LLM distillation.
    \item \textbf{User-Friendly Design}: KDFlow is a lightweight framework based on FSDP2, and decouples the algorithms from the whole distillation pipeline.
\end{itemize}

\section{Related Work}
\subsection{Knowledge Distillation for LLMs}
Knowledge Distillation (KD) was first proposed by \citet{hinton2015distilling} to compress large models into smaller ones. 
For LLMs, KD is commonly divided into black-box \cite{kim2016seqkd} and white-box distillation \cite{zhang2023tiekd,gu2023minillm,agarwal2024gkd,ko2024distillm,wu2025akl}. 
White-box KD aligns the student and teacher output distributions with divergence metrics, providing richer supervision and often better performance.
The paradigm has also expanded from off-policy distillation on static datasets to on-policy distillation, where the student learns from its own generated data \cite{gu2023minillm,agarwal2024gkd,xiao2026mimo}. 
Recent work further studies cross-tokenizer KD to handle vocabulary mismatches between teacher and student models \cite{wan2024fusellm,boizard2024uld,zhang2024dskd,zhang2025dskdv2,cui2025multilevelot,chen2025cdm,minixhofer2025cross}. 
Despite these algorithmic advances, a flexible and efficient framework for KD research remains lacking, which motivates this work.

\begin{figure*}
    \centering
    \includegraphics[width=0.75\linewidth]{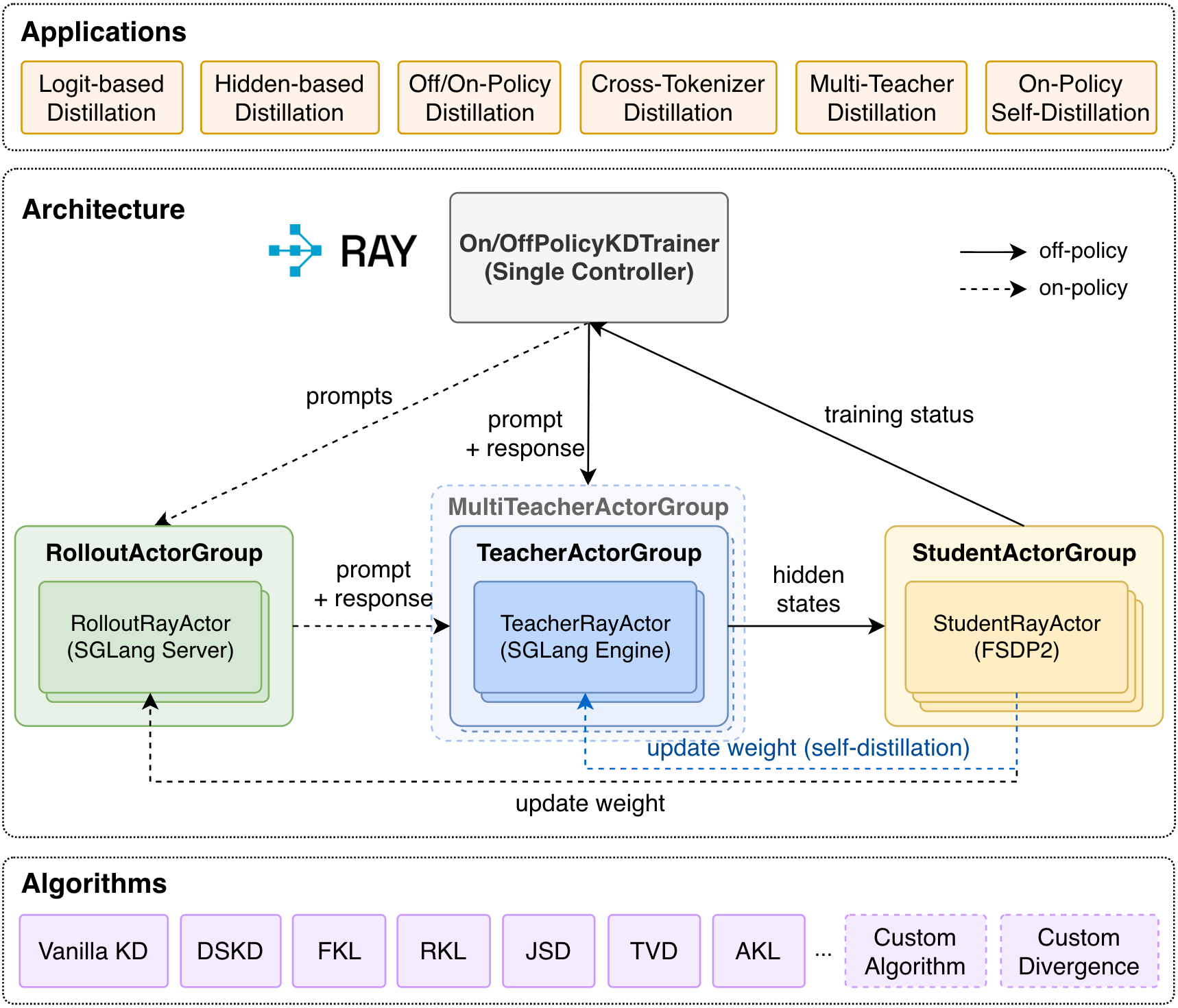}
    \caption{Overview of KDFlow. The framework is built on Ray \cite{moritz2018ray} and decouples the distillation pipeline by allocating the teacher model to SGLang and the student model to FSDP2. Solid and dashed arrows illustrate the data flow for off-policy and on-policy distillation, respectively. Notably, KDFlow transfers compact hidden states from the teacher rather than full logits to reduce communication overhead.}
    \label{fig:architecture}
\end{figure*}

\subsection{Existing Frameworks}
Existing LLM training and distillation frameworks provide useful support but still have important limitations. 
TRL \cite{vonwerra2020trl} and EasyDistill \cite{wang2025easydistill} support KD with homogeneous training engines, which underutilize hardware during teacher forward passes. 
Slime \cite{slime_github} and verl \cite{sheng2025hybridflow} decouple inference and training with engines like vLLM/SGLang, but they are not KD-specialized and only support on-policy distillation with incomplete logit information. 
MS-SWIFT \cite{zhao2025swift} supports both off-policy and on-policy distillation, but its implementation is relatively heavy.
By contrast, \textbf{KDFlow} is a lightweight and efficient KD framework covering full KD scenarios, as detailed in Table \ref{tab:framework_compare}.

\section{The Design of KDFlow}


The overall framework of KDFlow is presented in Figure \ref{fig:architecture}. To address the inefficiency of using a single homogeneous engine for both inference and training, we adopt a top-down decoupled design. In this section, we introduce the system architecture, the core communication mechanism, the distillation workflows, and the algorithm abstractions.

\subsection{System Architecture}

The KD process typically involves multiple stages: teacher inference, student training, and student rollout, which require different backends during training. 
Therefore, KDFlow is built upon Ray to manage distributed processes efficiently. 
As shown in the middle part of Figure \ref{fig:architecture}, the architecture consists of a single controller (Trainer) and three functionally independent actor groups:
\begin{itemize}
    \item \textbf{Trainer (Single Controller)}: Trainer is the central coordinator that manages the dataset, controls the training loop, and organizes the data flow among different actor groups. KDFlow supports both OffPolicyKDTrainer and OnPolicyKDTrainer.
    \item \textbf{RolloutActorGroup}: RolloutActorGroup is used for the rollout process of the student model during on-policy distillation. Following frontier RL frameworks like Slime \cite{slime_github}, it uses SGLang Router to connect multiple SGLang HTTP servers for load-balanced inference. Moreover, KDFlow uses a colocated mode and updates model weights in SGLang via CUDA Interprocess Communication (IPC).
    \item \textbf{TeacherActorGroup}: TeacherActorGroup manages multiple TeacherRayActor instances that execute the forward pass of the teacher. Specifically, we initialize an SGLang engine\footnote{The SGLang engine is more compatible with NumPy ndarray objects than the SGLang HTTP server.} in each TeacherRayActor to obtain the teacher's hidden states. This allows KDFlow to leverage SGLang's high-throughput and flexible parallel strategies for teacher-model inference.
    \item \textbf{StudentActorGroup}: StudentActorGroup is deployed with PyTorch FSDP2. It handles the standard training processes for the student model, including forward pass, backward pass, and optimizer state management. Then it returns training status (e.g., the loss values) for logging.
\end{itemize}
This decoupled architecture ensures that the inference-heavy teacher model and the training-heavy student model run on their respective optimized backends, significantly improving hardware utilization.

\begin{figure}[t]
    \centering
    \includegraphics[width=\linewidth]{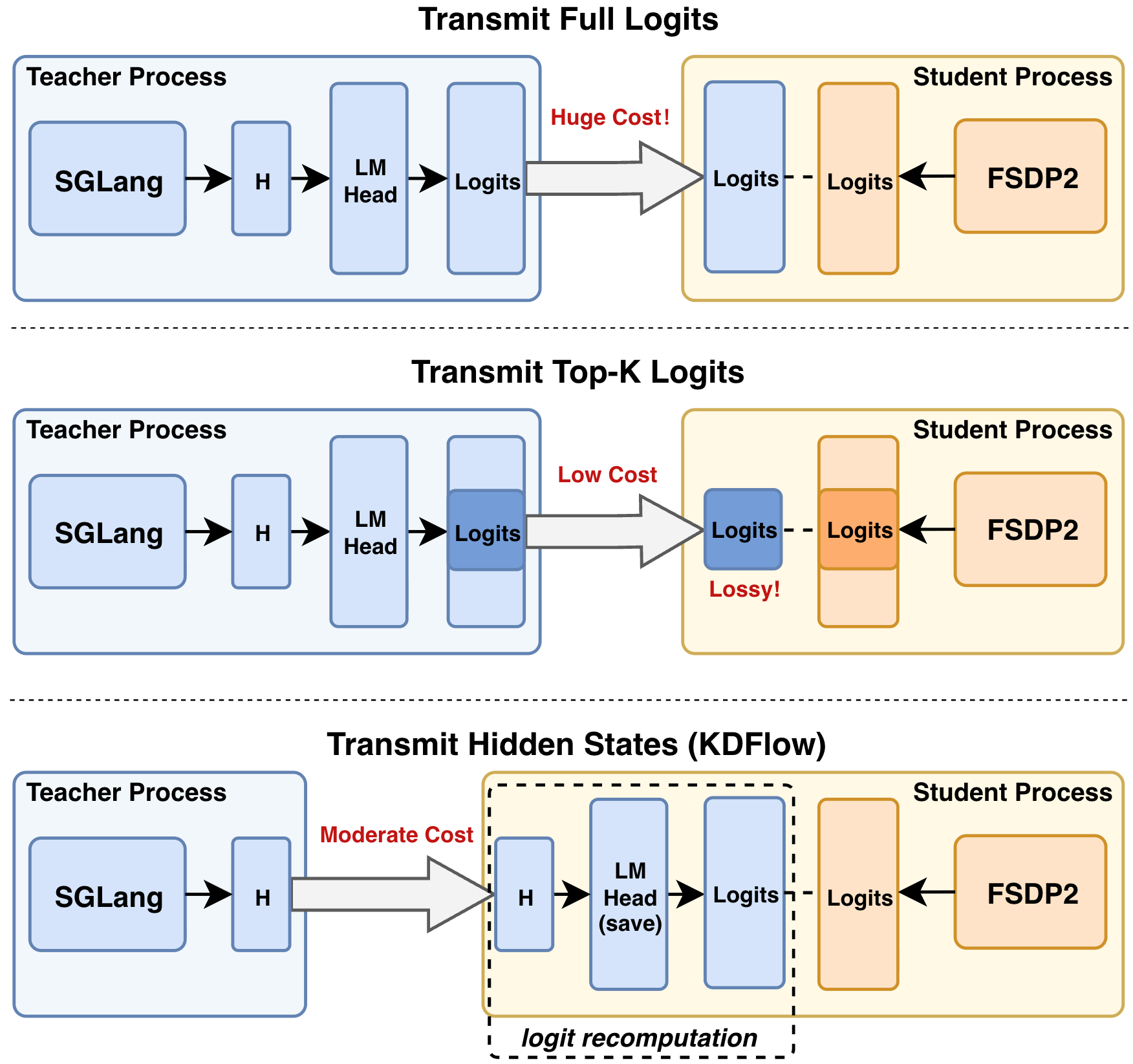}
    \caption{Comparison of different decoupled distillation approaches.}
    \label{fig:cost}
\end{figure}

\subsection{Efficient Communication via Hidden States}
In the decoupled architecture, the teacher and student typically run in separate processes, making the transfer of teacher knowledge a key bottleneck. 
Directly transmitting full logit distributions becomes prohibitive due to their large data volume, while transferring only top-k logits reduces bandwidth at the cost of breaking the mathematical equivalence of distillation and degrading performance.
To address this issue, KDFlow adopts hidden-state transfer and logit recomputation, as shown in Figure \ref{fig:cost}. 
Instead of sending full logits, the TeacherActorGroup outputs only the teacher's final hidden states, whose dimension (e.g., 4096) is much smaller than the vocabulary size (e.g., 151936).
KDFlow further uses shared memory and the Ray shared object mechanism for zero-copy transfer across processes.
After receiving the teacher's hidden states, each StudentRayActor locally recomputes the full logit distributions using the teacher's language model head. 
This design substantially reduces communication volume while preserving the mathematical equivalence to standard logit-based KD.

\subsection{Distillation Workflows}
Governed by the Single Controller, KDFlow seamlessly supports two main distillation workflows, illustrating the flexible data routing among the decoupled actors:

\textbf{Off-Policy Distillation} (Solid lines in Figure \ref{fig:architecture}): The student learns from a static dataset.
The Trainer sends the prompts and the corresponding responses directly to the TeacherActorGroup to obtain the teacher's hidden states. 
These hidden states, along with the inputs, are then passed to the StudentActorGroup to compute the distillation loss and update the student's weights.

\textbf{On-Policy Distillation} (Dashed lines in Figure \ref{fig:architecture}): The student learns from data generated by the student itself. 
First, the Trainer sends prompts to the RolloutActorGroup to generate responses. 
Next, these prompt-response pairs are sent to the TeacherActorGroup to obtain the teacher's hidden states. 
Then, the data and hidden states flow into the StudentActorGroup for gradient updates. 
Finally, the updated weights of the student model are synchronized back to the RolloutActorGroup to ensure the next generation step uses the latest policy.

\subsection{Comprehensive Abstractions and Algorithms}
To provide a user-friendly and out-of-the-box toolkit, KDFlow strictly separates the underlying system pipeline from the distillation algorithms.
As shown at the bottom of Figure \ref{fig:architecture}, KDFlow provides built-in support for various KD algorithms and divergence metrics, including Forward KL (FKL), Reverse KL (RKL), Jensen-Shannon Divergence (JSD), and Total Variation Distance (TVD). 
Users can also easily implement their custom distillation losses or algorithms with minimal code, eliminating the need to understand the complex distributed communication logic.
Furthermore, KDFlow natively supports cross-tokenizer distillation. 
When the teacher and student models have different vocabularies, directly aligning their full logit distributions is impossible. 
Therefore, KDFlow implements the DSKDv2 \cite{zhang2025dskdv2} algorithm for cross-tokenizer knowledge distillation.

With this decoupled design and efficient communication mechanism, KDFlow significantly improves the overall training throughput, which we will demonstrate in the following experiments.

\begin{figure}[t]
    \centering
    \includegraphics[width=\linewidth]{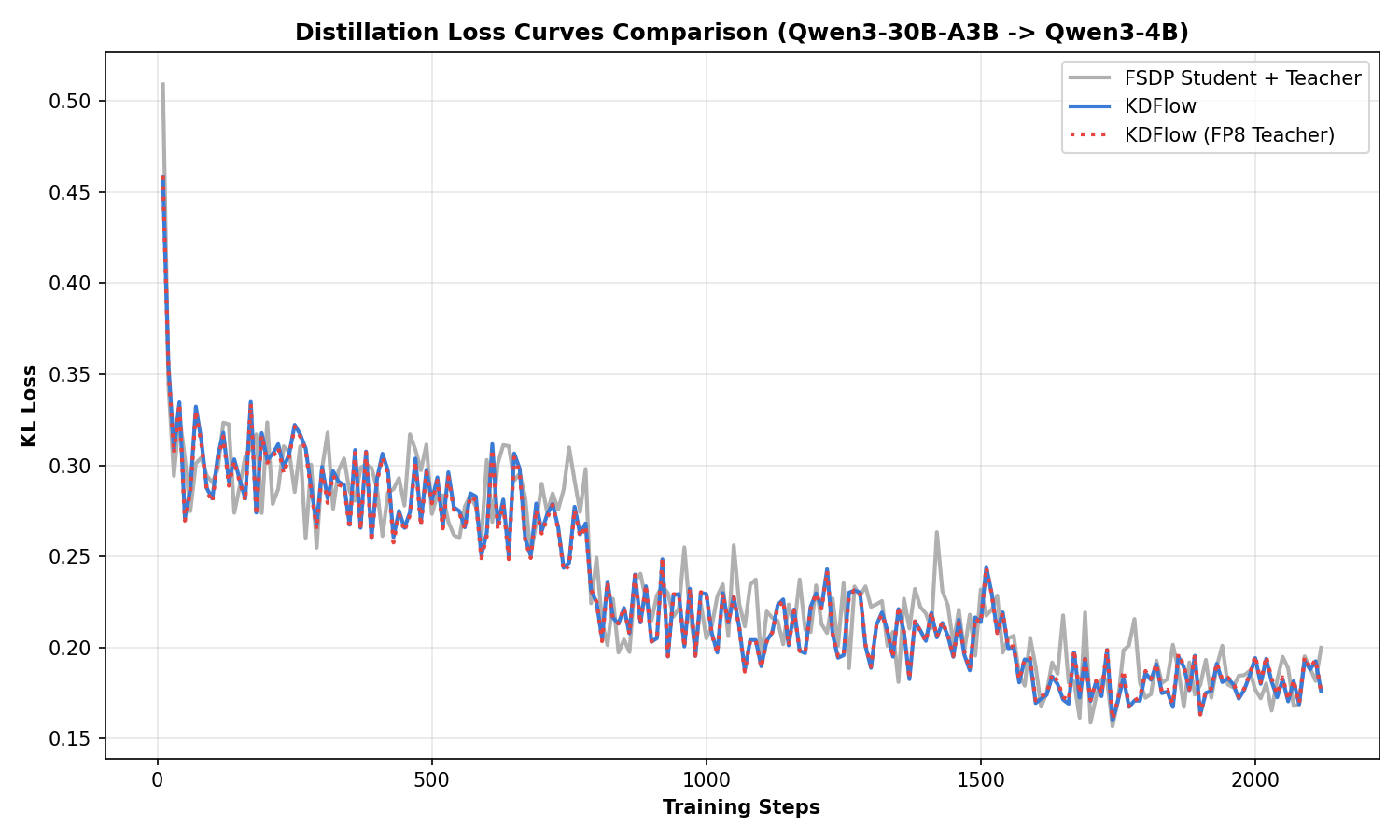}
    \caption{Loss curves of KDFlow and the pure FSDP implementation when distilling Qwen3-30B-A3B to Qwen3-4B.}
    \label{fig:loss_curves}
\end{figure}

\begin{table}[tbp]
    \centering
    \resizebox{\linewidth}{!}{
    \begin{tabular}{l l cc}
        \toprule
        \multirow{2}{*}{\textbf{Model}} & \multirow{2}{*}{\textbf{Framework}} & \multicolumn{2}{c}{\textbf{AlpacaEval 2.0}} \\
        \cmidrule(lr){3-4}
        & & \small{LC-Win Rate (\%)} & \small{Win Rate (\%)} \\
        \midrule
        Original Student & Qwen3-1.7B & 26.09 & 21.99 \\
        \midrule
        \multirow{4}{*}{Distilled Student} & MS-SWIFT & \textbf{28.40} & 27.86 \\
        & KDFlow (Pure FSDP) & 28.18 & 28.20 \\
        & KDFlow & 28.23 & 28.32 \\
        & KDFlow (On-Policy) & 28.29 & \textbf{29.95} \\
        \midrule
        Teacher & Qwen3-30B-A3B & 68.65 & 66.52 \\
        \bottomrule
    \end{tabular}
    }
    \caption{Student model performance on AlpacaEval 2.0 after distillation with different frameworks.}
    \label{tab:distill_performance}
\end{table}

\begin{table*}[htbp]
    \centering
    \resizebox{\textwidth}{!}{
    \begin{tabular}{l c ccc ccc}
        \toprule
        \multirow{2}{*}{\textbf{Framework}} & \multirow{2}{*}{\textbf{\makecell{Training \\Backend}}} & \multicolumn{3}{c}{\textbf{Student: Qwen3-4B}} & \multicolumn{3}{c}{\textbf{Student: Qwen3-1.7B}} \\
        \cmidrule(lr){3-5} \cmidrule(lr){6-8}
        & & Qwen3-14B & Qwen3-32B & Qwen3-30B-A3B & Qwen3-14B & Qwen3-32B & Qwen3-30B-A3B \\
        \midrule
        TRL & ZeRO-3           & 21.3s/it & 31.5s/it & - & 13.3s/it & 23.4s/it & - \\
        MS-SWIFT & ZeRO-3     & 16.6s/it & 24.8s/it & 43.2s/it & 11.5s/it & 20.1s/it & 36.9s/it \\
        ROLL & FSDP2         & 38.4s/it & 56.9s/it & 67.9s/it & 26.8s/it & 45.6s/it & 53.8s/it \\
        \hdashline
        KDFlow (BF16 Teacher) & FSDP2 & 12.3s/it & 15.7s/it & 11.3s/it & 7.6s/it & 10.9s/it & 5.9s/it \\
        \textbf{KDFlow (FP8 Teacher)}  & FSDP2 & \textbf{11.5s/it} & \textbf{13.5s/it} & \textbf{11.1s/it} & \textbf{6.7s/it} & \textbf{8.7s/it} & \textbf{5.8s/it} \\
        \midrule
        \textbf{Speedup} & - & \textcolor{green!50!black}{\textbf{1.44$\times$}} & \textcolor{green!50!black}{\textbf{1.84$\times$}} & \textcolor{green!50!black}{\textbf{3.89$\times$}} & \textcolor{green!50!black}{\textbf{1.72$\times$}} & \textcolor{green!50!black}{\textbf{2.31$\times$}} & \textcolor{green!50!black}{\textbf{6.36$\times$}} \\
        \bottomrule
    \end{tabular}
    }
    \caption{Off-policy training efficiency comparison (seconds per iteration) across different distillation frameworks. \textbf{Speedup} is calculated using KDFlow (FP8) against the best-performing baseline (i.e., MS-SWIFT). All frameworks use identical training settings: global batch size=128, gradient accumulation=8, and max length=4096.}
    \label{tab:training_speed}
\end{table*}

\section{Experiments}
In this section, we comprehensively evaluate the KDFlow framework from the following five perspectives: 
(1) \textbf{Loss Correctness:} We validate the loss correctness of KDFlow by recording and analyzing its loss curves against those of the standard FSDP baseline. 
(2) \textbf{KD Performance:} We compare the knowledge distillation performance of KDFlow with that of baseline frameworks on a representative downstream task.
(3) \textbf{Training Efficiency:} We compare KDFlow with existing frameworks in both off-policy and on-policy distillation settings.
(4) \textbf{Multi-Node Scalability:} We evaluate how KDFlow scales from one to eight nodes.
(5) \textbf{Cross-Tokenizer Distillation:} We evaluate KDFlow when the teacher and student use different tokenizers.

\subsection{Experimental Setup}

We evaluate KDFlow on both instruction-following and mathematical-reasoning distillation tasks.
The loss validation, downstream performance, and off-policy efficiency experiments use multiple Qwen3 student sizes with both dense and MoE teachers; the on-policy experiments additionally cover a DeepSeek-based configuration on DAPO-Math-17K, while the cross-tokenizer experiment distills Qwen3-14B into Llama3.2-3B.
Off-policy framework comparisons use a single 8-GPU server, the on-policy efficiency comparison uses two 8-GPU nodes, and the scalability experiment uses up to eight such nodes.
We report loss curves, AlpacaEval 2.0 performance in standard and cross-tokenizer settings, end-to-end training time, and multi-node scaling behavior.
Detailed model, dataset, baseline, hardware, and training configurations are provided in Appendix \ref{sec:experimental_setup_details}.

\subsection{Loss Curve Validation}

The core technical innovation of KDFlow is to leverage SGLang as the teacher backend and collect the teacher's hidden states to recompute full logits on the student side. 
However, recent studies on RL training have shown that high-throughput inference engines may introduce numerical discrepancies during model execution, particularly for MoE models \cite{yao2025tis,qi2025fp16,ma2025r3,zheng2025stabilizing}.
This motivates us to investigate whether such discrepancies affect the distillation behavior of KDFlow.
Specifically, we compare the loss curves produced by KDFlow with those of a standard KD implementation, where the teacher and student are executed within the same FSDP process. 
As shown in Figure \ref{fig:loss_curves}, the loss curves of KDFlow align well with the baseline throughout the entire training process. 
Moreover, the loss curves almost overlap for FP8 and BF16 teacher inference, which suggests that FP8 teacher inference is an efficient alternative for LLM distillation.

\subsection{Downstream Task Performance}

Beyond training loss, we evaluate the effectiveness of the student models trained via KDFlow on downstream tasks to ensure our decoupled architecture does not compromise distillation performance. 
We perform off-policy distillation from Qwen3-30B-A3B to Qwen3-1.7B using the Forward KL (FKL) divergence across all frameworks.
Table \ref{tab:distill_performance} presents the performance evaluation on the AlpacaEval 2.0 benchmark. 
The student model distilled via KDFlow achieves an LC-Win Rate of 28.23\% and a Win Rate of 28.32\%, demonstrating significant improvement over the un-distilled Qwen3-1.7B baseline. 
The on-policy variant further improves the Win Rate to 29.95\%, showing the flexibility of KDFlow across different distillation workflows.
Moreover, when compared to the students distilled using the MS-SWIFT and pure FSDP baselines, the performance differences remain within a negligible range. 
We also report the teacher model performance as a reference, where Qwen3-30B-A3B achieves an LC-Win Rate of 68.65\% and a Win Rate of 66.52\%.
These results indicate that KDFlow safely optimizes the system execution pipeline and reduces communication overhead while strictly preserving the distillation quality. 
KDFlow acts as a transparent, high-performance infrastructure that faithfully executes KD algorithms without sacrificing downstream performance.

\subsection{Training Efficiency}

\subsubsection{Off-Policy Distillation Efficiency}

We compare the training speed of KDFlow with existing LLM distillation frameworks in the off-policy setting, where the rollout cost in on-policy distillation is excluded to better isolate the efficiency of teacher-student training.
Specifically, we measure the average training time per step across different teacher-student setups.
As shown in Table \ref{tab:training_speed}, KDFlow is consistently the fastest framework among all FSDP2- and DeepSpeed ZeRO-3-based candidates, achieving a 1.44$\times$ to 6.36$\times$ speedup over MS-SWIFT.
The advantage is especially pronounced for the MoE teacher Qwen3-30B-A3B, where standard training engines suffer from inefficient routing and expert management.
By serving the teacher with SGLang and transferring compact hidden states instead of full logits, KDFlow reduces the distillation time to 5.8s/it.

\subsubsection{On-Policy Distillation Efficiency}
\label{subsubsec:on_policy_distillation_efficiency}
We further compare the end-to-end efficiency of KDFlow against verl and MS-SWIFT in the on-policy distillation setting on two nodes.
The experiments are conducted on DAPO-Math-17K \cite{yu2025dapo} with two student-teacher configurations: (1) Qwen3-1.7B-Base as the student and Qwen3-14B as the teacher, and (2) DeepSeek-R1-Distill-1.5B as the student and JustRL-DeepSeek-1.5B as the teacher.
Detailed hyperparameters and deployment configurations are provided in Appendix \ref{sec:experimental_setup_details}.

Table \ref{tab:on_policy_efficiency} reports end-to-end epoch time, covering rollout generation, teacher inference, student optimization, and weight synchronization.
Under the same 16-GPU budget, verl partitions the GPUs between the student and teacher, whereas KDFlow colocates both models across the shared GPU pool. This design allows GPUs to be reused across alternating inference and training stages, reducing idle capacity and improving overall resource utilization.
Overall, KDFlow achieves the lowest epoch time in both configurations: 8.1 and 3.7 hours, corresponding to 1.75$\times$ and 1.43$\times$ speedups over verl, respectively, demonstrating consistent efficiency gains across different student-teacher pairs.

\begin{table}[t]
    \centering
    \setlength{\tabcolsep}{4pt}
    \renewcommand{\arraystretch}{1.15}
    \resizebox{\linewidth}{!}{
    \begin{tabular}{lcccccc}
        \toprule
        \multirow{2}{*}{\textbf{Framework}}
        & \multirow{2}{*}{\makecell{\textbf{OPD} \\ \textbf{Style}}}
        & \multirow{2}{*}{\makecell{\textbf{GPU} \\ \textbf{Allocation}}}
        & \multicolumn{2}{c}{\makecell{\textbf{Qwen3-14B} $\rightarrow$ \\ \textbf{Qwen3-1.7B-Base}}}
        & \multicolumn{2}{c}{\makecell{\textbf{JustRL-1.5B} $\rightarrow$ \\ \textbf{DS-R1-Distill-1.5B}}} \\
        \cmidrule(lr){4-5} \cmidrule(lr){6-7}
        & & & \textbf{Time (h)} & \textbf{Speedup} & \textbf{Time (h)} & \textbf{Speedup} \\
        \midrule
        verl & PG & 8S + 8T & 14.2 & 1.00$\times$ & 5.3 & 1.00$\times$ \\
        \multirow{2}{*}{MS-SWIFT} & PG & 8S + 8T & 18.6 & 0.76$\times$ & 13.7 & 0.39$\times$ \\
        & GKD & Colocate & 10.6 & 1.34$\times$ & 5.8 & 0.91$\times$ \\
        \textbf{KDFlow} & GKD & Colocate & \textbf{8.1} & \textcolor{green!50!black}{\textbf{1.75$\times$}} & \textbf{3.7} & \textcolor{green!50!black}{\textbf{1.43$\times$}} \\
        \bottomrule
    \end{tabular}
    }
    \caption{End-to-end time per epoch for on-policy distillation on DAPO-Math-17K. Speedup is calculated relative to verl in each setting. All frameworks use FSDP2 for student training. ``8S + 8T'' indicates 8 GPUs for the student and 8 GPUs for the teacher.}
    \label{tab:on_policy_efficiency}
\end{table}

\subsection{Multi-Node Scalability}

We evaluate the scalability of KDFlow using the same two on-policy configurations as in \S \ref{subsubsec:on_policy_distillation_efficiency} on 1, 2, 4, and 8 nodes, each with eight NVIDIA H20 GPUs, while fixing the dataset, training configuration, and one-epoch workload.

\begin{figure}[t]
    \centering
    \includegraphics[width=\linewidth]{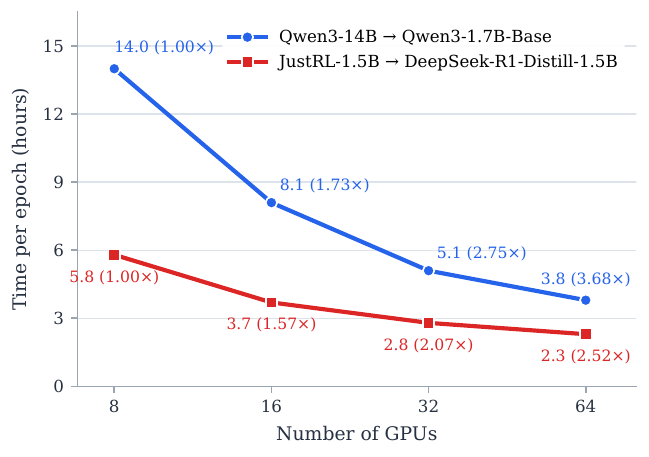}
    \caption{Multi-node scalability of KDFlow for on-policy distillation. Scaling from 8 to 64 GPUs yields a 3.68$\times$ speedup for the Qwen3 setting and a 2.52$\times$ speedup for the DeepSeek setting. Each node contains eight NVIDIA H20 GPUs.}
    \label{fig:multi_node_scalability}
\end{figure}

As shown in \ref{fig:multi_node_scalability}, KDFlow consistently benefits from additional computational resources across both configurations. For the Qwen3 configuration, increasing the number of GPUs from 8 to 64 reduces the epoch time from 14.0 hours to 3.8 hours, yielding a 3.68$\times$ speedup. For the DeepSeek configuration, the epoch time decreases from 5.8 hours to 2.3 hours, corresponding to a 2.52$\times$ speedup. The Qwen3 configuration exhibits stronger scaling, as its larger teacher model benefits more from the parallel computation of additional GPU resources. Overall, these results demonstrate that KDFlow effectively scales beyond a single node and can support large-scale on-policy distillation across distributed GPU clusters.

\subsection{Cross-Tokenizer Distillation}

We further verify KDFlow in a cross-tokenizer setting, where the teacher and student use different tokenizers and cannot be directly aligned through vanilla full-vocabulary KL divergence.
Specifically, we distill knowledge from Qwen3-14B into Llama3.2-3B and find that KDFlow's DSKD-based objective further improves over sequence-level KD.
As shown in Table \ref{tab:cross_tokenizer_performance}, DSKD + KL improves the LC-Win Rate from 29.72\% to 32.88\% and the Win Rate from 39.32\% to 42.54\% compared with sequence-level KD.

\begin{table}[t]
    \centering
    \resizebox{\linewidth}{!}{
    \begin{tabular}{l l cc}
        \toprule
        \multirow{2}{*}{\textbf{Model}} & \multirow{2}{*}{\textbf{Setting}} & \multicolumn{2}{c}{\textbf{AlpacaEval 2.0}} \\
        \cmidrule(lr){3-4}
        & & \small{LC-Win Rate (\%)} & \small{Win Rate (\%)} \\
        \midrule
        Original Student & Llama3.2-3B & 15.26 & 18.45 \\
        \midrule
        \multirow{2}{*}{Distilled Student} & SeqKD & 29.72 & 39.32 \\
        & DSKD + KL & \textbf{32.88} & \textbf{42.54} \\
        \midrule
        Teacher & Qwen3-14B & 73.31 & 72.51 \\
        \bottomrule
    \end{tabular}
    }
    \caption{Cross-tokenizer distillation performance on AlpacaEval 2.0.}
    \label{tab:cross_tokenizer_performance}
\end{table}

\section{Conclusion}
In this paper, we present \textbf{KDFlow}, an efficient and user-friendly framework for LLM knowledge distillation, which decouples SGLang-based teacher inference from FSDP2-based student training. 
By transferring teacher hidden states and recomputing logits on the student side, KDFlow reduces communication overhead while preserving theoretical equivalence to full-vocabulary KD.
KDFlow achieves 1.44$\times$ to 6.36$\times$ speedups over MS-SWIFT in off-policy distillation and 1.43$\times$ to 1.75$\times$ speedups over verl in on-policy distillation, while demonstrating strong multi-node scalability.
With native support for off-policy, on-policy, and cross-tokenizer distillation, KDFlow provides a flexible infrastructure for efficient LLM compression and post-training.

\section*{Limitations}

While KDFlow significantly improves the efficiency and flexibility of LLM distillation, it has certain limitations.
First, although KDFlow demonstrates effective scaling up to 64 GPUs, its current student training backend is built entirely upon PyTorch FSDP2 and lacks the complex 3D parallelism of Megatron-LM \cite{shoeybi2019megatron}; scalability to substantially larger models and clusters therefore remains unverified.
Second, as a research-oriented framework, KDFlow currently lacks some industrial-grade optimizations, such as asynchronous training, which are crucial for training on clusters with thousands of GPUs.
However, KDFlow explicitly prioritizes the needs of the research community: user-friendliness, high flexibility, and rapid prototyping. 
By abstracting away complex distributed communication logic, KDFlow allows researchers to easily implement and test novel off-policy, on-policy, or cross-tokenizer distillation algorithms with minimal engineering overhead. 
Integrating Megatron-LM and further optimizations remain important directions for our future work.



\bibliography{custom}

\newpage
\appendix

\section{Experimental Setup Details}
\label{sec:experimental_setup_details}

\subsection{Off-Policy Distillation}

\paragraph{Models and Data.}
We randomly sample 100,000 instruction-following prompts from LMSys-Chat-1M \cite{zheng2023lmsyschat1m} and generate responses with Qwen3-14B.
The loss validation, downstream performance, and efficiency experiments use Qwen3-14B, Qwen3-32B, and Qwen3-30B-A3B as teachers and Qwen3-4B and Qwen3-1.7B as students \cite{yang2025qwen3}.
The cross-tokenizer experiment uses Qwen3-14B as the teacher and Llama3.2-3B as the student.
For AlpacaEval 2.0, we use Qwen3-235B-2507-Instruct as the evaluator model.

\paragraph{Baselines and Configuration.}
We compare KDFlow with TRL \cite{vonwerra2020trl}, ROLL \cite{wang2025roll}, and MS-SWIFT \cite{zhao2025swift}.
The efficiency experiments use a global batch size of 128, gradient accumulation of 8, and a maximum sequence length of 4,096.
All off-policy experiments run on one server with 8 NVIDIA H20 GPUs and CUDA 12.9.

\subsection{On-Policy Distillation}

\paragraph{Models and Data.}
We use DAPO-Math-17K \cite{yu2025dapo} and evaluate two teacher--student configurations: Qwen3-14B $\rightarrow$ Qwen3-1.7B-Base and JustRL-DeepSeek-1.5B $\rightarrow$ DeepSeek-R1-Distill-1.5B.

\paragraph{Training Configuration.}
The maximum input and generation lengths are 1,024 and 7,168, respectively; each rollout batch contains 64 prompts and four sampled responses per prompt.
All frameworks use the same dataset and generation settings, with optimization hyperparameters matched where supported.

\paragraph{Baselines and Hardware.}
We compare KDFlow with verl \cite{sheng2025hybridflow} and MS-SWIFT.
The efficiency comparison uses two identical 8-GPU H20 nodes with CUDA 12.9: verl and MS-SWIFT with PG-OPD allocate 8 GPUs each to the student and teacher, whereas MS-SWIFT with GKD and KDFlow colocate them over the shared 16 GPUs.
The scalability experiments use the same one-epoch workloads and generation settings on one, two, four, and eight identical nodes.




\end{document}